\pdfoutput=1
\documentclass[11pt]{article}

\usepackage[]{ACL2023}

\usepackage{times}
\usepackage{latexsym}

\usepackage[T1]{fontenc}
\usepackage[utf8]{inputenc}

\usepackage{microtype}

\usepackage{inconsolata}
\usepackage{booktabs} 

\usepackage{graphicx}
\usepackage{subfigure}
\usepackage{parskip}
\usepackage{xcolor}
\usepackage{multirow}
\usepackage{url}
\usepackage{mdframed}
\RequirePackage{rotating}

\title{Semantic Differentiation in Speech Emotion Recognition: Insights from Descriptive and Expressive Speech Roles}

\author{
  Rongchen Guo$^1$\thanks{ \quad Equal contributions},  Vincent Francoeur$^2$\footnotemark[1], Isar Nejadgholi$^{1,3}$\footnotemark[1], \\ \textbf{Sylvain Gagnon$^2$, Miodrag Bolic$^1$\thanks{\quad Corresponding author}}\\
  $^1$School of Electrical Engineering and Computer Science, University of Ottawa \\
  $^2$School of Psychology, University of Ottawa \\
  $^3$National Research Council Canada, Ottawa, Canada \\
  \texttt{\{rongchen.guo, vfran022, sgagnon, Miodrag.Bolic\}@uottawa.ca}\\
  \texttt{isar.nejadgholi@nrc-cnrc.gc.ca}
}

\begin{document}
\maketitle

\begin{abstract}

% Speech Emotion Recognition (SER) has emerged as a crucial field for enhancing human-computer interaction, yet its accuracy remains constrained by the complexity of emotional nuances in speech. In this study, we propose a novel framework that differentiates two distinct semantic roles: \textit{descriptive} semantics, which relate to the contextual content of speech, and \textit{expressive} semantics, which reflect the speaker's emotional state. Leveraging a carefully curated dataset of emotionally evocative movie clips, we collected intended emotion tags of movies and audiences' ratings of actual evoked emotions, along with valence and arousal scores. A three-step methodology was employed: automatic speech recognition, semantic segmentation, and emotion prediction. Our findings demonstrate that descriptive semantics predominantly align with intended emotions, while expressive semantics correlate more strongly with evoked emotions, offering new insights into semantic roles in SER. This differentiation enhances classification and regression performance, advancing the understanding and application of emotion-based AI systems, such as customer service, mental health support, and human-AI interaction. By highlighting the interplay between semantics and emotion, this work lays a foundation for future SER research and practical implementations in personalized and context-aware technologies.

Speech Emotion Recognition (SER) is essential for improving human-computer interaction, yet its accuracy remains constrained by the complexity of emotional nuances in speech. In this study, we distinguish between \textit{descriptive semantics}, which represents the contextual content of speech, and \textit{expressive semantics}, which reflects the speaker's emotional state. After watching emotionally charged movie segments, we recorded audio clips of participants describing their experiences, along with the intended emotion tags for each clip, participants' self-rated emotional responses, and their valence/arousal scores. Through experiments, we show that descriptive semantics align with intended emotions, while expressive semantics correlate with evoked emotions.
%, leading to improved classification and regression performance. 
Our findings inform SER applications in %customer service, mental health support, and 
human-AI interaction and pave the way for more %personalized and 
context-aware AI systems.

\end{abstract}

\begin{figure*}[t]
    \centering
\includegraphics[width=\linewidth]{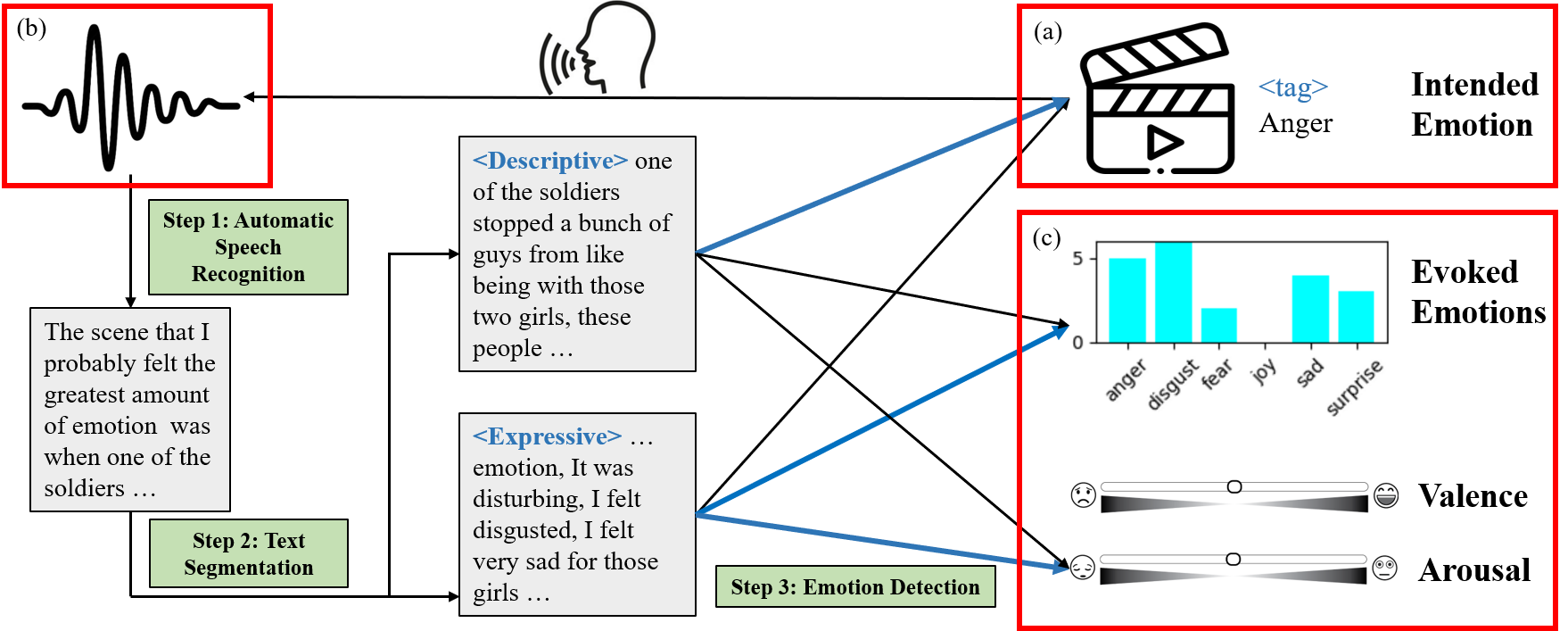}
\caption{\textbf{Data Collection and Algorithm Workflow:}  Participants watched six videos eliciting specific emotions and provided speech descriptions, emotion ratings, and valence/arousal scores. Speech data were transcribed, segmented into \textbf{descriptive} and \textbf{expressive} semantics, and used to train models for three tasks: predicting intended emotions (TASK-1), evoked emotions (TASK-2), and valence/arousal (TASK-3).}
%\caption{\textbf{Data Collection and Algorithm Workflow}. Each participant was asked to watch six videos with an intended emotion (e.g. anger) to be elicited (Figure~\ref{fig:flow}-a). We collected three types of data for each movie segment from the participants: (1) a 30-second speech describing their feelings towards the video (Figure~\ref{fig:flow}-b), (2) participants' emotional ratings on a scale of 1 to 6 for each of the six emotions (Figure~\ref{fig:flow}-c), and (3) the reported overall emotional valence and arousal (Figure~\ref{fig:flow}-c). 
%We designed three tasks to explore the role of speech semantics, given the speech data collected in Figure~\ref{fig:flow}-b; \textbf{TASK-1):} predict the intended emotion associated with the given movie in Figure~\ref{fig:flow}-a, \textbf{TASK-2):} predict the actual evoked emotions in Figure~\ref{fig:flow}-c, and \textbf{TASK-3):} predict the participants rated emotional valence and arousal in Figure~\ref{fig:flow}-c.
    % To achieve this, we followed a three-step framework: the speech data were first transcribed into texts using an ASR system. Then, the speech transcriptions were segmented into \textit{descriptive} semantics and \textit{expressive} semantics according to the semantic roles. Finally, classifiers/regressors were trained to detect the speech emotions.
   % }
    \label{fig:flow}
\end{figure*}

\section{Introduction}

The ability to accurately detect and interpret emotions in speech is vital for developing intelligent systems capable of natural and empathetic human-computer interactions. Speech Emotion Recognition (SER) has gained significant traction in recent years, driven by applications ranging from virtual assistants to mental health monitoring~\cite{ley2019evaluating,Rumpa2015Validating}. 
Despite these advancements, SER faces persistent challenges due to the complex and multi-dimensional nature of emotions, which often intertwine with contextual and speaker-specific factors.

Traditional approaches to SER have largely focused on acoustic features, such as pitch, energy, and spectral properties, to infer emotional states~\cite{wu2011automatic,bitouk2010class,venkataramanan2019emotion,likitha2017speech,kwon2003emotion}. While effective to some extent, these methods often overlook the semantic content of speech, which can provide crucial contextual information. 
With the advances in natural language processing, it has become increasingly feasible to analyze the semantic aspects of speech for emotion recognition~\cite{tzirakis2021speech, xu2021exploring}.
%Recent advances in natural language processing have enabled researchers to explore semantic roles in SER, 
However, the interplay between semantic roles and emotional expression remains underexplored. 
Specifically, the distinction between \textit{intended emotions} elicited by a stimulus and \textit{evoked emotions} experienced by the speaker is rarely addressed, leaving a critical gap in the field.

This paper introduces a novel framework to address this gap by distinguishing two types of semantic roles in speech. We hypothesize that 
\textit{Descriptive semantics} captures scenario-specific content, such as the narrative or context described in the speech. In contrast,
\textit{Expressive semantics} reflects the speaker’s subjective emotional stance, shaped by their personal experiences and delivery style.
In our framework, descriptive segments are expected to align with the intended emotion of the stimulus (the target emotion the video was designed to elicit), while expressive segments are expected to align with the evoked emotion (the participant’s self-reported experience). This mapping allows us to distinguish stimulus-driven affect from speaker-specific affect, thereby addressing a critical gap in prior SER research that often assumes a single ground-truth label.
%By separating these semantic roles, our approach enables a deeper understanding of the relationship between linguistic content and emotional expression. 
This semantic distinction is particularly important in settings where it is essential to understand not only \textit{what happened} — the contextual content of speech — but also \textit{how it was felt} — the speaker’s emotional state and tone.
Such an understanding has practical implications for applications like emotion-aware AI systems, educational tools, and interactive entertainment, where both the content and emotional delivery of speech play key roles in creating engaging and effective human-computer interactions.

To validate our hypothesis, we collected a dataset comprising emotionally evocative movie clips to elicit a specific emotion.
Participants watched the videos and provided ratings for the actual evoked emotions, alongside valence and arousal scores, creating a robust foundation for analysis.
Our methodology to uncover the distinct relationships between semantic roles and intended versus evoked emotions involves three key steps: speech transcription with automatic speech recognition (ASR), semantic segmentation with LLMs, and emotion prediction with fine-tuned text classifiers/regressors. 
This work makes the following contributions: 
\begin{itemize}
    
\item First, we curated a SER dataset with 582 audio recordings spanning six emotion categories. Audio transcriptions are generated, and intended emotions, as well as evoked emotions, are measured in an experimental setup. 
% \footnote{We will share the anonymized  transcripts and ratings with the research community, based on research agreement and upon request.}  

\item Second, we implemented an LLM-based semantic segmentation approach to separate the expressive and descriptive parts of speech and validated that through human evaluation. 

\item Third, through experimentation, we show that descriptive semantics are more predictive of intended emotions, while expressive semantics are better aligned with evoked emotions. %Our findings can inform future research and experiment design in SER, with potential applications in virtual assistants, customer service, and mental health support.
\end{itemize}

Importantly, our work goes beyond simply predicting emotion labels from participants’ descriptions. By explicitly segmenting speech into descriptive and expressive roles, we quantify how different semantic functions relate to stimulus-intended versus self-experienced emotions. This role-based separation provides a principled way to reconcile discrepancies between intended and evoked affect and offers interpretable insights that are not available from standard text-only or audio-only models.

Our findings have significant implications for designing more accurate and context-aware emotion recognition systems, with potential applications in virtual assistants, customer service, and mental health support. By bridging the gap between semantics and emotion, this research advances the state-of-the-art in SER and sets the stage for future exploration of semantic roles in emotional AI systems.

\section{Related Work}

Emotions are complex psychological and physiological responses to salient events, involving bodily sensations, expressive behaviors, and cognitive evaluations~\cite{moors2024overview,moors2009theories}. Various linguistic features, including prosody, lexical choice, and sentence structure, play a role in the perception and expression of emotions~\cite{mohammad2010emotions,barrett2007experience,keltner2019emotional}.
Speech emotion recognition (SER) models aim to detect emotional states from speech using acoustic, textual, or multimodal signals. 
With the advancement of LLMs and automatic speech recognitions (ASR), text-based emotion classification has seen improved accuracy~\cite{hama2024emotion, bekmanova2022emotional, bharti2022text}. 
Acoustic-based emotion detectors have also progressed using acoustic feature extractors, such as openSMILE~\cite{eyben2010opensmile} or audio embedding models, such as wav2vec~\cite{baevski2020wav2vec} and HuBERT~\cite{hsu2021hubert}, which embed paralinguistic cues such as pitch, tempo, and energy into speech representations~\cite{ulgen2024revealing, chakhtouna2024unveiling, zhao2024knowledge, dutta2024leveraging, ghosh2016representation}.
Multi-modal approaches, which combine speech, facial expressions, and physiological signals, have also become increasingly prominent in recent years~\cite{cheng2024emotion, khan2024mser, morency2017multimodal, yoon2018multimodal, niu2023capturing}.

Emotion elicitation via multimedia stimuli (e.g. short film clips) is a common technique in SER to induce targeted emotions (e.g., sadness, joy, fear)~\cite{Li2021Effectiveness,Rumpa2015Validating}.
% Machine learning algorithms utilizing facial feature tracking and physiological data have shown promise in real-time emotion classification \cite{Bailenson2008Real}. 
These movie-based emotion elicitation techniques have applications in various fields, including e-health monitoring and human-computer interaction~\cite{ley2019evaluating,Rumpa2015Validating}.
The stimuli are selected and validated through self-report and physiological measures~\cite{Chen2021Selection,Handayani2015Recognition,Soleymani2012Multimodal}.
While these methods control for the \textit{intended} emotional target, they do not always account for the \textit{evoked} emotion the speaker experiences and expresses. 
Prior work such as ~\citet{siedlecka2019experimental} reviewed these paradigms in detail, but focused primarily on affect induction rather than the emotional content of participants' verbal responses. In our work, we analyze speech collected after stimulus exposure and study how intended and evoked emotions are reflected in participants’ spoken descriptions. In doing so, we explore a novel distinction between semantic roles in language—namely, whether a speaker is being descriptive (e.g., summarizing the movie) or expressive (e.g., conveying their own reaction)—and how these roles align with different emotion types.

Many SER datasets have been developed. 
In acted speech datasets, such as IEMOCAP~\cite{busso2008iemocap} and SAVEE~\cite{jackson2014surrey}, 
actors are recruited to read sentences or act in scenes that portray different emotions.
In spontaneous speech datasets, such as MSP-Podcast~\cite{lotfian2017building},
MSP-Conversation~\cite{martinez2020msp},
SAMAINE~\cite{mckeown2011semaine},
and RECOLA~\cite{ringeval2013introducing},
and elicited speech datasets, such as LSSED~\cite{fan2021lssed}, 
BAUM-1~\cite{zhalehpour2016baum}, 
and eNTERFACE~\cite{batlinerreleasing},
audios are recorded in a freely speaking environment or with emotion elicitation methods. Speech is then annotated by a third party (perception-of-other).
However, these datasets focus on one emotion label per speech and do not distinguish different types of emotions.
% Most commonly, MSP-Podcast~\cite{lotfian2017building},
% MSP-Conversation~\cite{martinez2020msp},
% LSSED~\cite{fan2021lssed}, 
% and other works collect speech annotated by a third party (perception-of-other). 
% Several existing datasets have addressed related goals but from different perspectives. 
% Most other commonly used SER datasets have only one emotion label per speech and do not distinguish different types of emotions, such as intended emotional target and the speaker's own emotional states.
To this end, EMO-DB~\cite{burkhardt2005database} and IEMOCAP~\cite{busso2008iemocap} analyzed emotional evocative sentences and perception-of-other in acted speech. 
Most similar to us, MuSE~\cite{jaiswal2020muse} collects speech following emotional video stimuli and reports both self-reported and intended emotion annotations. While similar in structure, our work uniquely interprets the relationship between stimulus-intended and self-reported emotions through a \textit{semantic} lens,
% Other commonly used SER datasets~\cite{lotfian2017building, martinez2020msp, busso2008iemocap, busso2016msp} often emphasize either acted emotional expressions or third-party emotion perception (perception-of-other), rather than focusing on the speaker's own emotional states. 
% By contrast, our study captures both the intended emotional target and the speaker's self-reported emotion, 
enabling direct analysis of misalignment between the two emotion types.

Furthermore, some recent studies in NLP have explored emotion elicitation and manipulation in conversational settings~\cite{gong2023eliciting, ma2025detecting, qian2023think, meng2024revisiting}.  While our study does not model conversational interactions, our semantic framework may offer insights into these settings by helping to identify when emotional influence is being attempted or received. For example, expressive speech segments may signal internal affective states, while descriptive segments may reflect contextual awareness or narrative framing. These distinctions could inform models of emotion transfer and regulation in human-computer dialogue.

Our contribution lies in bridging the gap between stimulus-based emotion elicitation and the actual emotions conveyed by participants in speech. By segmenting utterances according to their semantic roles and analyzing how different roles align with either intended or evoked emotions, we propose a novel way to interpret emotional speech beyond traditional modality-based or label-based approaches.
While prior SER studies have emphasized either acoustic or multimodal representations, our work suggests that semantic structure in language - accessible only through text - offers a distinct and interpretable signal for differentiating between types of emotion.

% When detecting emotions from speech data, researchers leverage acoustic features and textual features for emotion detection. The advent of LLMs and improved ASR systems have enhanced performance in text-based emotion detection. Multi-modal databases incorporating facial expressions, voice, and physiological signals further advance the field \cite{Soleymani2012Multimodal}.

% Despite progress, challenges remain in standardizing emotion elicitation methods and ensuring their validity \cite{Jemioo2021Emotion}. In this work, we explore the variability in semantic content within verbally expressed emotions to better understand their relationship with the intended and evoked emotions. Specifically, we aim to examine %how different semantic elements correlate with emotional responses to deepen insights into 
% the nuanced connections between targeted emotional stimuli and participants' actual emotional experiences.

\begin{table*}[t]
\small
\centering
\begin{tabular}{p{3.5cm}lp{5cm}ll}
\hline
\multicolumn{1}{c}{\textbf{Movie Clip}} & \multicolumn{1}{c}{\textbf{Tag}} & \multicolumn{1}{c}{\textbf{Scene Description}}                                                      & \multicolumn{1}{c}{\textbf{Duration}} & \multicolumn{1}{c}{\textbf{Validation Source}}         \\ \hline
The Blair Witch Project~\cite{myrick1999blair}                 & Fear                                          & Final scene when screaming intensifies, man standing facing the wall and camera falls.              & 2:03                                  & \citet{schaefer2010assessing} \\
The Conjuring~\cite{conjuring2013}                           & Fear                                          & Girl gets out of bed at night and bags her head on a cupboard. Frantic scene.                       & 2:26                                  & \citet{iyilikci2024extended} \\
American History X~\cite{kaye1998american}                     & Anger                                         & Neo-Nazi kills a black man, smashing his head on the curb and then smiles after being arrested.     & 3:24                                  & \citet{schaefer2010assessing} \\
Platoon~\cite{stone1986platoon}                                 & Anger                                         & Villagers pushed around in burning village and soldier stops other soldiers from raping a child.    & 2:42                                  & Author tested in pilot. \\
Baby laughing at ripping paper~\cite{youtube2011baby}          & Joy                                           & 8-month-old Micah (a boy) laughing hysterically while at-home daddy rips up a job rejection letter. & 1:44                                  & Author tested in pilot. \\
Cats and Dog playing together~\cite{youtube2011mom}           & Joy                                           & Dog lies peacefully on a large bed with kittens and adult cat moving around. With happy music.      & 1:53                                  & Author tested in pilot. \\
One Day~\cite{scherfig2011one}                                 & Surprise                                      & Woman rides a bicycle; she gets hit by a truck.                                                     & 2:26                                  & \citet{zupan2020eliciting} \\
Neighbors~\cite{stoller2014neighbors}                               & Surprise                                      & Woman calls man about missing airbags Man is ejected to an office ceiling.                          & 1:07                                  & Author tested in pilot. \\
Trainspotting ~\cite{boyle1996trainspotting}                          & Disgust                                       & The main character enters “The worst toilet in Scotland” and later dives into a filthy toilet bowl. & 1:23                                  & \citet{schaefer2010assessing} \\
Planet Terror~\cite{rodriguez2007planet}                           & Disgust                                       & Scene where man is examined by doctors in a hospital and exposes infected parts of his body.        & 2:01                                  & \citet{michelini2019latemo} \\
Young impala and dead mother~\cite{youtube2022saddest}            & Sadness                                       & Young impala finds adult impala lying down and apparently dead. Then lies by dead animal.           & 1:44                                  & Author tested in pilot. \\
My Girl ~\cite{zieff1991my}                                & Sadness                                       & Funeral scene where girl cries and runs away after approaching the casket where a little boy lies.  & 1:39                                  & \citet{gabert2015ratings} \\ \hline
\end{tabular}
\caption{Listing and information about the 12 movie clips used to elicit discrete emotions in the main study.}
\label{tab:movie-info}
\end{table*}

\section{Dataset}
\label{sec:data}
The block diagram in Figure~\ref{fig:flow} summarizes our data collection, task definitions, and methodology, which we will elaborate on here and in the next section. Data collection was carried out in person at INSPIRE Laboratory of the School of Psychology at University of Ottawa. The experiment procedure was approved by the Research Ethics Board of University of Ottawa.
%A total of 99 students participated in the study. Data from two participants were excluded due to limited English proficiency, resulting in a final sample of 97 participants. The included participants ranged in age from 18 to 27 years (M = 19.9, SD = 2.5). The majority were women (81 women, 15 men, and 1 non-binary), and most were native English speakers (65 English, 12 French, and 20 other languages). 
The study included 97 student participants aged 18 to 27 (M = 19.9, SD = 2.5). The majority were women (81 women, 15 men, and 1 non-binary), and most participants were native English speakers (65 spoke English as their first language, 12 spoke French, and 20 spoke other languages). The sample was ethnically diverse, comprising 16 Asian, 20 Black/African, 7 Hispanic/Latino, 1 Indigenous, 15 Mixed/Multiple Ethnicities, 33 White/Caucasian, and 5 participants identifying as Other.

Our study focused on the six basic emotions identified by ~\citet{ekman1992facial} as the target emotions in our experimental setup: sadness, fear, joy, disgust, surprise, and anger. Two movie clips for each emotion were sourced from film stimuli in the existing literature and validated in our pilot study. 
% (see Appendix~\ref{appx:movie} for the movie sources and descriptions). 
The twelve movie clips used in the study and their meta-information are listed in Table~\ref{tab:movie-info}. 
We trimmed clips to ensure optimal emotional salience and duration. Their effectiveness was validated in a pilot study with 25 participants before the final data collection.

%Overall, 97$\times$ 6 audio clips were collected in the dataset, as well as the rated emotions with discrete scores converted to a scale from 0 to 6, and valence and arousal with continuous scores on a scale from 0 to 1, as rated by participants.

In the main study, participants watched six emotional video clips, one from each emotion category. To re-establish baseline levels of valence and arousal, the presentation of each emotional clip was preceded by a neutral video clip. To further mitigate potential carryover effects between conditions, a two-minute rest period was inserted between each neutral–emotional clip sequence, during which one of six still images was displayed on the computer screen. All video clips and still images were presented in random order to minimize potential sequence effects. 
% with a neutral clip shown between each to allow their emotions to return to baseline before proceeding to the next emotional clip. %The movie clips were selected based on their demonstrated ability to elicit specific emotional responses (six emotions, two clips per emotion). 
The collected dataset consists of 97$\times$ 6 entries, with five elements: 
1) \textbf{Speech}: a 30-second audio recording of the participant's verbal response to the following instruction: \textit{"You are asked to verbally describe the scene during which you felt the strongest emotion in the last film clip and say how it made you feel."}
2) \textbf{Intended emotions}: Each video is expected to provoke a certain emotion.  
3) \textbf{Evoked emotions}: the intensities at which each of the emotions (sadness, fear, joy, disgust, surprise, anger) was felt, as rated by the participants on a 7-point Likert scale going from \textit{not at all} to \textit{strongly}. 
4) \textbf{Valence}: the extent to which the overall feeling of the participant was positive or negative. 
5) \textbf{Arousal}: the intensity of the overall feeling of the participant while watching the video. Valence and arousal were measured on a validated sliding scale where each extreme was illustrated by an emoticon.

%An example data from one participant after watching one particular movie segment is shown 

%The block diagram in Figure~\ref{fig:flow} summarizes our data collection, task definitions, and methodology. We will elaborate further on tasks and methodology in the following sections. 

\section{Tasks}
\label{sec:task}

We define three tasks to examine the relationship between semantic types and emotion recognition. To determine the most predictive semantic type for each task, we experimented with three different inputs: full transcriptions, descriptive semantic segments, and expressive semantic segments.% (1) classification of intended emotions, (2) classification of evoked emotions, and (3) regression of emotion valence and arousal. 
%Each task captures a distinct aspect of how semantics influence emotion representation.

\textbf{TASK-1: Classification of Intended Emotion} involves classifying the intended emotion associated with each video based on participants' speech. %Intended emotions %represent the affective state the video is designed to evoke (e.g., sadness, joy, fear) and 
%are categorized into six classes: sadness, fear, joy, disgust, surprise, and anger.
%Comparing performance across these inputs allows us to determine the most predictive semantic type.

\textbf{TASK-2: Classification of Evoked Emotion} involves classifying participant-reported evoked emotions, which are subjective and may include multiple emotions simultaneously. While evoked emotions often include the intended emotion, individual differences can lead to variations. This task is framed as a multi-label classification problem, where each emotion (on a scale of 0 to 6) is binarized based on whether it is evoked or not. %The predictive power of expressive segments reflects their alignment with speaker-specific emotional experiences.

\textbf{TASK-3: Regression of Valence and Arousal} predicts participants' self-reported valence and arousal ratings, which provide a two-dimensional representation of emotional states. %Valence measures emotional positivity or negativity, while arousal represents intensity. Both are rated on a continuous scale from 0 to 1. 

\section{Methodology}

As depicted in Figure \ref{fig:flow}, our %proposed framework integrates semantic analysis into the SER pipeline to enable a deeper understanding of the role of speech content in emotion detection. The 
methodology consists of three sequential steps: speech recognition, semantic segmentation, and emotion prediction. 
% Implementation details are provided in appendix \ref{implementation}.

% We distinguish two semantic roles:

% The framework works in three steps: First, speech data is transcribed into texts with an ASR model. Then, a semantic segmentator is adopted to extract descriptive semantics and expressive semantics from the transcription. Lastly, classifiers and regressors are trained on T1, T2, and T3 based on different semantics as inputs.

% The hypothesis is that \textit{descriptive} speech semantics are more correlated with intended emotion detection tasks (T1) and \textit{expressive} speech semantics are more correlated with evoked emotion detection tasks (T2 \& T3). Figure.~\ref{fig:flow} illustrates the overall workflow, with details of each component provided in the subsections below.

\textbf{Step-1: Automatic Speech Recognition - }
We used Whisper~\cite{radford2023robust}, an automatic speech recognition model, to transcribe the participants' speech data %are first transcribed 
into text. %with an automatic speech recognition model. 
% In this work, we use Whisper~\cite{radford2023robust}, a state-of-the-art system known for its robustness across diverse accents and noise conditions. 
% To ensure accuracy, we manually reviewed all transcriptions and corrected any minor errors.

\textbf{Step-2: Semantic Segmentation - }
%We defined two types of speech semantics: descriptive semantics and expressive semantics.
%Descriptive segments provide factual or narrative information about the video content, while expressive segments convey personal emotions and subjective reactions. 
We used GPT-4o ~\cite{gpt4o} %as the text segmentator 
to extract descriptive %segments 
and expressive segments from the transcription obtained in step 1. 
The prompt is given in Table~\ref{tab:prompt}. %The text segmentator can be either a prompt-based LLM or a fine-tuned model specific for text segmentation task.
We set the sampling temperature to 0 to make the process more deterministic. 
Overlapping phrases were allowed when semantic roles intersected, ensuring comprehensive representation.

% We prompted GPT-4o\footnote{\url{https://platform.openai.com/docs/models/gpt-4o}} to perform this semantic segmentation task. The specific prompt used for segmentation is provided below. To ensure accuracy, the segmentation results were reviewed and refined by the authors. Overlapping phrases were allowed when semantic roles intersected, ensuring comprehensive representation.

% \begin{mdframed}
% The user will provide a paragraph describing their feelings towards a particular movie, delimited with ```\#\#\#\#```. 

% Your task is to segment the paragraph into two parts according to the type of content: descriptive segments and expressive segments.

% Descriptive segments refer to elements or clauses that provide factual or narrative information about the movie content without explicitly reflecting personal emotions or opinions.

% Expressive segments refer to elements or clauses that convey personal feelings, attitudes, or opinions. These segments reflect individual reactions, emotions, and perceptions, or the intensity of these emotions.

% The two parts (descriptive segments and expressive segments) can overlap, but all clauses of the given paragraph must be contained in at least one of the two parts.

% Output your answer in the following format:

% <answer>

%   ~~~~<descriptive>[descriptive segments]</descriptive>
  
%   ~~~~<expressive>[expressive segments]</expressive>
  
% </answer>
% \end{mdframed}

\textbf{Step-3: Emotion Prediction - }
The last step is to %predict the emotions with different semantic segments as inputs.
%To test our hypothesis, we design a set of emotion classification / regression 
perform tasks described in Section~\ref{sec:task} to study the relationship between semantic roles and emotion recognition.
Each model is trained and evaluated on three input types: full transcriptions, descriptive segments, and expressive segments. 
% Models are described in appendix \ref{implementation}.

 %We assess the predictive power of each semantic type to detect intended and invoked emotion by comparing the performance of models in different tasks.

% In the last step, we perform the classification of intended emotions (T1) and evoked emotions (T2), and regression of emotion valence and arousal (T3), based on the semantic contents extracted in the previous step.

% A BERT model~\cite{devlin2018bert} was trained on the segmented transcripts to perform the three predefined tasks. The text segments were input into BERT, with the $[CLS]$ token embedding from the final layer representing the entire segment. A linear layer was then added on top of this embedding for downstream classification. The only difference between the model architectures across the three tasks lies in the training objectives. For Task 1 (T1), a multi-class classification task, we used cross-entropy loss. For Task 2 (T2), a multi-label classification task, we applied binary cross-entropy loss with a sigmoid layer. For Task 3 (T3), a regression task, we used mean squared error loss.

\textit{Audio-Based Emotion Classification - }
In addition to text-based models, we also experimented with audio-based models trained directly on the speech recordings. These included a HuBERT model~\cite{hsu2021hubert}, a Wav2Vec2 model~\cite{baevski2020wav2vec}, and a baseline MFCC (mel-frequency cepstral features) classifier. The audio classifiers were evaluated on TASK-1 and TASK-2 using the full utterance audio as input. However, all speech-based models performed significantly worse than text-based classifiers. Since semantic role segmentation (i.e., distinguishing between descriptive and expressive segments) is inherently a linguistic task and not inferable from acoustic signals alone, we prioritized text-based methods for the core analyses of this paper moving forward.

\begin{table}[t]
\small
\begin{tabular}{lcc}
\hline
                           & Descriptive & Expressive
                         \\& semantics & semantics
                         \\ \hline
LLM \& Annotator 1         & 0.71        & 0.73       \\
LLM \& Annotator 2         & 0.84        & 0.83       \\
Annotator 1 \& Annotator 2 & 0.77        & 0.74       \\
Random \& Random           & 0.63        & 0.64       \\ \hline
\end{tabular}
\caption{Human evaluations of GPT-4o text segmentations. The agreement between two human annotators was comparable to human-LLM agreements.}
\label{tab:human-eval}
\end{table}

\begin{table}[t]
\small
    \centering
    \begin{tabular}{|p{7.2cm}|}
        \hline
        The user will provide a paragraph describing their feelings towards a particular movie, delimited with ```\#\#\#\#```. \\\\
Your task is to segment the paragraph into two parts according to the type of content: descriptive segments and expressive segments.
\\\\
Descriptive segments refer to elements or clauses that provide factual or narrative information about the movie content without explicitly reflecting personal emotions or opinions.
\\\\
Expressive segments refer to elements or clauses that convey personal feelings, attitudes, or opinions. These segments reflect individual reactions, emotions, and perceptions, or the intensity of these emotions.
\\\\
The two parts (descriptive segments and expressive segments) can overlap, but all clauses of the given paragraph must be contained in at least one of the two parts.
\\\\
Output your answer in the following format:
\\\\
<answer>
\\
  ~~<descriptive> [descriptive segments] </descriptive>
 \\ 
  ~~<expressive> [expressive segments] </expressive>
\\  
</answer>\\\hline
    \end{tabular}
    \caption{Prompt for extracting descriptive and expressive semantics from speech transcription.}
    \label{tab:prompt}
\end{table}

\begin{table*}[t]
\centering
\begin{tabular}{p{13cm}c}
\hline
How often do participants experience the intended emotion conveyed by the videos? & 96.63\% \\
How frequently do participants feel emotions other than the intended one?         & 89.39\% \\
How often is the intended emotion rated as the highest by participants?     & 79.29\% \\ 
Chippendale's alpha coefficient%~\cite{krippendorff2011computing, krippendorff2018content} 
 between intended emotion and evoked emotions %(see Appendix~\ref{alpha} for details)     
& 0.1466 \\ 
\hline
\end{tabular}
\caption{Statistics of relationships between movie intended emotion tags and evoked emotions. Predicting the evoked emotions is a much more subjective task than predicting the intended emotion tag.}
\label{tab:data-stat}
\end{table*}

\begin{figure*}[t]
    \centering
\includegraphics[width=\linewidth]{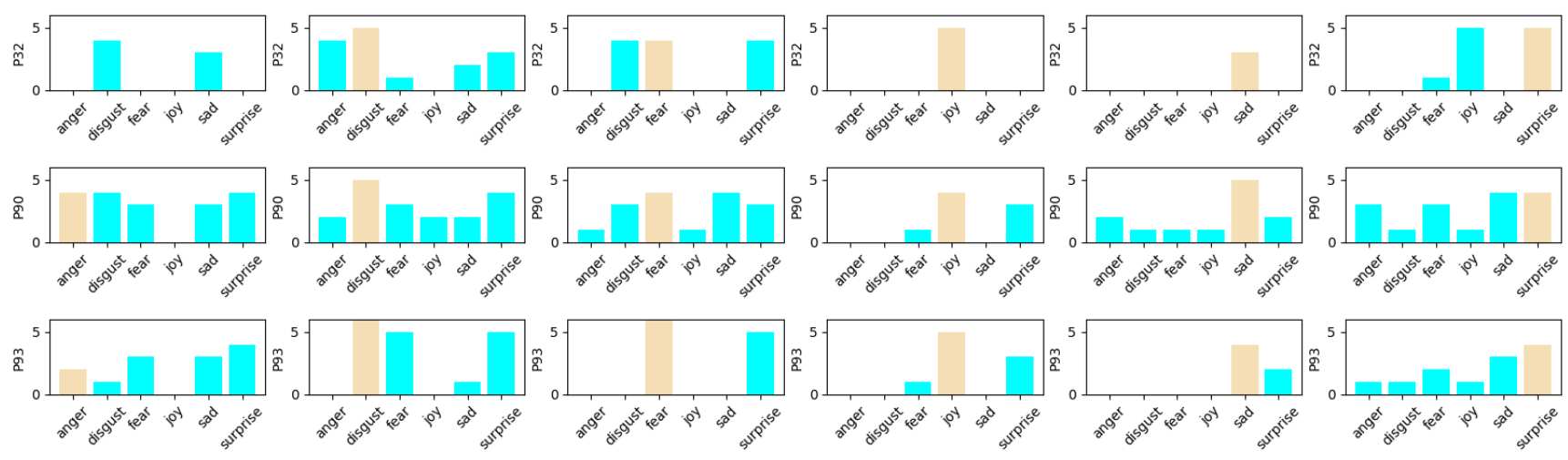}
    \caption{Examples of participants' rated emotions. Each row represents a participant who watched six movie segments (6 columns) from each of the six emotional categories. The intended emotion tag associated with the video is plotted in a yellow bar. Other rated emotions are colored blue. The height of the bars represents the emotion ratings from participants. 
    For example, in the second movie clip watched by Participant P93, the intended emotion was "disgust," as shown by the yellow bar. After watching the clip, P93 reported experiencing four emotions: disgust, fear, sadness, and surprise, indicated by the blue bars. Among these, ``disgust'' was the strongest emotion, receiving the highest score of 6. 
    }
    \label{fig:t2demo}
\end{figure*}

\section{Experiments}

\subsection{Implementation Details}

For Step-1, automatic speech recognition, we used `whisper-large-V3`~\footnote{https://huggingface.co/openai/whisper-large-v3}, a state-of-the-art system ASR model known for its robustness across diverse accents and noise conditions. We manually reviewed transcriptions in the development set, consisting of 33 $\times$ 6 audio transcriptions. Whisper achieved a 4.13\% word error rate, with errors mainly in unclear utterances at speech boundaries and between clauses.
 
For Step-2, we validated the effectiveness of GPT-4o segmentation again on the development set with 33 $\times$ 6 transcriptions. 
Two authors of this paper, one from the Computer Science department and the other from the Psychology department, were given the same instructions as the LLM and independently performed the same segmentation task. 
% These two human annotators were considered blind to the LLM segmentation results. 
To calculate the agreement between human annotators and also between LLM and annotators, we
computed cosine similarities of the segments, using sentence-transformer embeddings~\footnote{https://huggingface.co/sentence-transformers/all-MiniLM-L6-v2}. From Table~\ref{tab:human-eval}, the average agreement between two human annotators (0.76) was comparable to human-LLM agreement (0.73 and 0.83). Most discrepancies arose from minor conjunctions to make sentences more complete. 
As a baseline, two random text segmentations would result in a similarity score of 0.63 - 0.64. Overall, GPT-4o has an acceptable segmentation quality.

For Step-3, emotion prediction, we fine-tuned different classifiers/regressors, including BERT~\cite{devlin2018bert}, RoBERTa~\cite{liu2019roberta}, and DeBERTa~\cite{he2020deberta}. Different text semantics identified in Step-2 are used as inputs to the models.
For emotion classification (Tasks 1 and 2), we used the text embeddings from the models and applied a standard classification head with a softmax activation function to predict categorical emotions. For regression (Task 3), we modified the models by replacing the classification head with a fully connected layer that outputs a single continuous value, trained with mean squared error (MSE) loss to predict valence and arousal scores.
This approach follows standard practice in adapting transformer encoders for regression tasks~\cite{xin-etal-2021-berxit, taha2024text, orso2008bert}.
Data are split on participants' level, with 1/3 of participants (33 participants) data used for training, 1/3 for validation, and the rest 1/3 for testing. 
% Training and testing were done on a Tesla T4 GPU machine.

% \subsection{Implementation Details}

% We split the data on participants' level. 1/3 of participants (33 participants) data were used for training the models, 1/3 for validation, and the rest 1/3 for testing. 

% Experiments were done on a Tesla T4 GPU machine.
% %\subsection{Model Selection} 
% In Step-1, automatic speech recognition, we used Whisper~\cite{radford2023robust}, a state-of-the-art system known for its robustness across diverse accents and noise conditions. % To ensure accuracy, the authors manually reviewed all transcriptions and corrected any minor errors.
% For Step-2, we used GPT-4o ~\cite{gpt4o} as the text segmentator to extract descriptive and expressive semantics from speech transcriptions. We set the sampling temperature to 0 to make the process more deterministic. The prompt is given in Table~\ref{tab:prompt}.
% % To ensure accuracy, the segmentation results were reviewed and refined by the authors. 
% Overlapping phrases were allowed when semantic roles intersected, ensuring comprehensive representation.

% In Step-3, emotion prediction, we fine-tuned different classifiers/regressors, including BERT~\cite{devlin2018bert}, RoBERTa~\cite{liu2019roberta}, and DeBERTa~\cite{he2020deberta}.

%\subsection{Results}

\subsection{Comparative Analysis of Intended and Evoked Emotions} Table \ref{tab:data-stat} shows the relationship between the intended and evoked emotions. While the participants experienced the intended emotion 96.63\% of the time,  they also reported other emotions 89.39\% of the time. Surprisingly, more than 20\% of the time, an emotion other than the intended one is experienced most. These results suggest that the experienced emotion is highly subjective and can deviate from the intended emotions.

To better quantify the subjectivity of Task-2, we calculate the Krippendorff's alpha coefficient~\cite{krippendorff2011computing, krippendorff2018content} between the intended emotion and evoked emotions.
We treat the agreement between intended emotion and evoked emotion as the agreement 
% between two coders performing unit tests.
between two annotators performing multi-label annotations.
Each annotator labels 594 data points, since there are 99 $\times$ 6 speech.
One annotator always label the intended emotion as \textit{true} and other emotions as \textit{false}.
The other annotator labels the data with the participant's evoked emotion ratings in a multi-label fashion.
Krippendorff's alpha coefficient is calculated as the inter-annotator agreement index on this multi-label annotation task with MASI distance~\cite{passonneau2006measuring} as the distance measurement between two sets of multi-label annotations. The low score of Krippendorff's alpha coefficient shows the high subjectivity of task T2 and the high variation of evoked emotions with respect to the intended emotion.

Figure \ref{fig:t2demo} gives examples of emotion ratings by three different participants in response to the six movie segments. Each row in the grid represents data from a different participant, while each column corresponds to one of the six movie segments. Within the bar charts, yellow bars indicate the intended emotion that the video clip aimed to elicit, while blue bars represent the emotions self-reported by the participants after watching the clips. The height of the bars reflects the intensity of the rated emotions on a numerical scale. 
These examples highlight variability in participants' emotional responses, often revealing discrepancies between the intended emotions and the emotions participants actually experienced.

\begin{table*}[t]
\centering
% \small
\begin{tabular}{lcccc}
\hline
\textbf{Model} & \textbf{Semantics} & \textbf{Precision} & \textbf{Recall} & \textbf{F1} \\ \hline

BERT    & Descriptive  & 0.83  & 0.81  & 0.81  \\
        & Expressive   & 0.68  & 0.65  & 0.65  \\
        & Full        & 0.89  & 0.88  & 0.88  \\ \hline
RoBERTa & Descriptive  & 0.85  & 0.83  & 0.83  \\
        & Expressive   & 0.69  & 0.69  & 0.69  \\
        & Full        & 0.93  & 0.93  & 0.93  \\ \hline
DeBERTa & Descriptive  & 0.81  & 0.81  & 0.81  \\
        & Expressive   & 0.65  & 0.64  & 0.64  \\
        & Full        & 0.91  & 0.90  & 0.90  \\ \hline
\end{tabular}
\caption{Model performances on classifying intended emotion associated with the movies.}
\label{tab:task1}
\end{table*}

\begin{table*}[t]
% \small
\centering
\begin{tabular}{lcccc}
\hline
\textbf{Model} & \textbf{Semantics} & \textbf{Precision} & \textbf{Recall} & \textbf{F1} \\ \hline
BERT    & Descriptive  & 0.72   & 0.77   & 0.73   \\
        & Expressive   & 0.75  & 0.77   & 0.75   \\
        & Full        & 0.78   & 0.82   & 0.78  \\ \hline
RoBERTa & Descriptive  & 0.73   & 0.76   & 0.73   \\
        & Expressive   & 0.74   & 0.82  & 0.77   \\
        & Full        & 0.76   & 0.82   & 0.77   \\ \hline
DeBERTa & Descriptive  & 0.71   & 0.76   & 0.72   \\
        & Expressive   & 0.74   & 0.81   & 0.76   \\
        & Full        & 0.76   & 0.81   & 0.77   \\ \hline
\end{tabular}
\caption{Average model performances on classifying evoked emotions (std is always less than 0.1 over 5 run).}
\label{tab:task2}
\end{table*}

\begin{figure}[t]
    \centering
    \includegraphics[width=\linewidth]{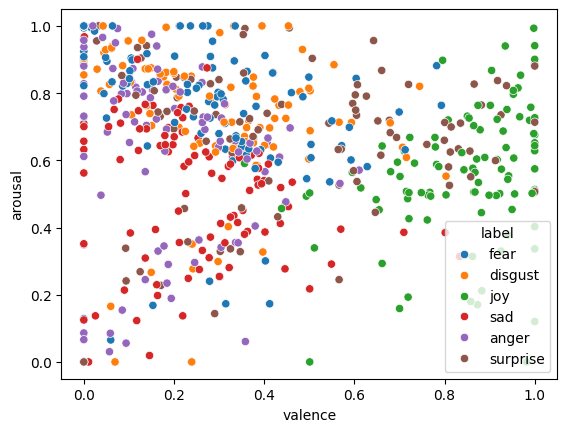}
    \caption{Valence and arousal ratings, colored by the intended emotion tags of movie segments.}
    \label{fig:t3demo}
\end{figure}

\begin{table*}[t]
\centering
% \small
\begin{tabular}{lcccccc}
\hline
\textbf{Model} & \textbf{Semantics} & \textbf{Valence MSE} & \textbf{Valence MAE} & \textbf{Arousal MSE} & \textbf{Arousal MAE} \\ \hline
BERT    & Descriptive  & 0.068  & 0.209  & 0.057  & 0.192  \\
        & Expressive   & 0.055  & 0.183  & 0.054  & 0.187  \\
        & Full        & 0.053  & 0.185  & 0.053  & 0.184  \\ \hline
RoBERTa & Descriptive  & 0.050  & 0.184  & 0.055  & 0.184  \\
        & Expressive   & 0.037  & 0.151  & 0.051  & 0.182  \\
        & Full        & 0.034  & 0.146  & 0.051  & 0.182  \\ \hline
DeBERTa & Descriptive  & 0.077  & 0.224  & 0.053  & 0.183  \\
        & Expressive   & 0.037  & 0.153  & 0.045  & 0.166  \\
        & Full        & 0.059  & 0.192  & 0.049  & 0.172  \\ \hline
\end{tabular}
\caption{Model performances on regression of emotion valence and arousal. Expressive semantics leads to smaller errors in estimating evoked valence and arousal. The difference is most pronounced for the DeBERTa-based model.}
\label{tab:task3}
\end{table*}

\begin{table*}[t]
\centering
\begin{tabular}{ccccccl}
\hline
         & \multicolumn{2}{c}{Valence} & \multicolumn{2}{c}{Arousal}                          \\
Model   & Z      & p                  & Z           & p                  \\ \hline
BERT    & $-1.74^a$ & $0.083$              & $-1.24^a$      & $0.215$             \\
RoBERTa  & $-2.98^a$ & $0.003$             & $-0.16^a$      & $0.874$              \\
DeBERTa & $-5.86^a$ & $<0.001$ & $-3.70^a$      & $<0.001$ \\ \hline
\end{tabular}
\caption{Wilcoxon signed-rank tests results to compare MSE between descriptive and expressive semantics for each model. 
% Smaller errors from predictions of valence using expressive semantics were statistically significant for both RoBERTa and DeBERTa models. Smaller errors from predictions of arousal with expressive semantics were significant for DeBERTa only. 
$^a$Based on positive ranks.}
\label{wilcoxon}
\end{table*}

\subsection{Classification of Intended Emotion} Aligned with our hypothesis, the classification results for TASK-1 demonstrate a clear advantage of using descriptive semantics as input for predicting the intended emotions associated with each movie segment. Table ~\ref{tab:task1} shows the classification accuracy for both semantic types across three different classifiers.
%: BERT, RoBERTa, and DeBERTa. 
Across all models, descriptive semantics consistently yield significantly higher accuracy in predicting the intended emotions. %This observation aligns with the hypothesis that descriptive semantics, which capture the contextual and narrative content of speech, provide a stronger foundation for aligning with the emotion explicitly intended by the movie clips.

%These results highlight the importance of descriptive semantics in capturing the content-driven aspects of speech, which are more closely aligned with the targeted emotional outcomes in controlled scenarios like movie viewing.

\subsection{Classification of Evoked Emotion} In TASK-2, we classified participant-reported evoked emotions, which are inherently subjective and may include multiple emotions simultaneously. Aligning with our hypothesis that expressive semantics better capture speaker-specific emotional experiences, results in Table~\ref{tab:task2} indicate that using expressive semantics as input achieves higher classification accuracy compared to using descriptive semantics. We also observe that even with full semantics, TASK-2 achieves significantly lower F-scores than Task-1, as expected due to the subjectivity of this task. 

%These findings highlight the utility of expressive semantics in capturing subjective emotional nuances and provide a foundation for future SER models to emphasize semantic roles tailored to speaker-specific emotional states.

\subsection{Discussion of Audio-Based Classifications}
% We also trained audio-only classifiers for TASK-1 and TASK-2 to assess the effectiveness of acoustic features in emotion recognition. The HuBERT model~\footnote{facebook/hubert-base-ls960} achieved 27.4\% and 21.5\% accuracy on TASK-1 and TASK-2 respectively, while Wav2Vec2~\footnote{facebook/wav2vec2-base} yielded 23.6\% and 19.6\%. A baseline MFCC model, which used mel-frequency cepstral features and a simple neural classifier, performed the worst with accuracies of 18.3\% and 16.4\%, often predicting only a single dominant class. In contrast, the text-based classifiers consistently achieved over 60\% accuracies on TASK-1 and 50\% accuracies on TASK-2. These results suggest that the prosodic and paralinguistic cues in our dataset were not strongly indicative of emotion, potentially due to participants’ steady and emotionally neutral delivery. Furthermore, these acoustic models do not offer a clear pathway to segment utterances by semantic roles, which limits their interpretability for our central research questions on distinguishing descriptive against expressive semantics.
To assess the role of acoustic features in emotion recognition, we trained several audio-only classifiers, including models based on HuBERT~\footnote{facebook/hubert-base-ls960}, Wav2Vec2~\footnote{facebook/wav2vec2-base}, and MFCC features, for both TASK-1 and TASK-2. Across all models, we observed consistently poor performance, with classifiers frequently defaulting to one or two majority emotion classes. This suggests that prosodic and paralinguistic cues in our dataset were not strongly indicative of emotional content. One likely explanation is that participants generally delivered their responses in a steady and emotionally neutral tone, which limited the expressiveness of acoustic features.

Moreover, unlike text-based inputs, speech signals do not easily lend themselves to semantic segmentation without speech recognition~\cite{wang2003speech, ong2005semantic}. Audio-based classifiers cannot distinguish between descriptive and expressive segments in an obvious way, making it difficult to explore the semantic roles that are central to our research questions. While acoustic features are valuable in many speech emotion recognition tasks, in our study design where subjective emotional experience is linked to semantic framing, textual cues proved more informative and interpretable.

\subsection{Regression of Emotion Valence and Arousal:}

Figure~\ref{fig:t3demo} shows the distributions of valence and arousal across different intended emotions, which  exhibit high variability without clear patterns across emotions. Positive emotions, such as joy, correlates with higher valence, and negative emotions, such as fear, have lower valence and higher arousal. But there is no obvious clusters among the six emotions. 

The results reported in Table~\ref{tab:task3} show that expressive semantics lead to more accurate predictions for both emotional valence and arousal compared to descriptive semantics.
A statistical analysis in Table~\ref{wilcoxon} shows that the differences in the prediction errors between descriptive and expressive semantics are statistically significant for valence under two of the three models and one model for arousal. 
The regression results are in line with the TASK-2 results and the statistical analysis partially supports the hypothesis that expressive semantics better capture subjective experience.

\section{Conclusion}

This study introduces a novel framework for Speech Emotion Recognition (SER) by distinguishing between  semantic roles in speech. By leveraging LLMs' zero-shot capabilities in text segmentation, we tackle a previously difficult challenge. To our knowledge, this is the first work to segment speech into two semantic roles, \textit{expressive} and \textit{descriptive} content, to enable more fine-grained and nuanced emotion detection. %By leveraging the zero-shot capabilities of large language models (LLMs), in semantic segmentation of text, we address a challenge that was previously much harder to tackle without such advanced tools.

% This study examines the role of speech semantics in SER. 
Our findings reveal that descriptive semantics are more predictive of intended emotions, while expressive semantics are more closely aligned with evoked emotions and their valence and arousal dimensions. This differentiation can inform future research in emotion detection. In some contexts, it might be more useful to instruct users and guide them toward only one of these modes of expressing emotions. In other applications, it might be more suitable to leave it to the users to express their emotions in a mixture of expressive and descriptive modes. The LLMs can then be used to segment the speech and use the segments depending on the predictive goals. This approach enhances the development of more accurate and context-aware emotion recognition systems, with applications in mental health, virtual assistants, and customer service. %provides deeper insights into the interplay between semantics and emotion, offering a new perspective on how linguistic content contributes to emotional expression.

% Our findings highlight the value of semantic segmentation in advancing SER methodologies and understanding emotion dynamics.

%The implications of this work extend beyond SER. Researchers can design experiments more intentionally by eliciting descriptive or expressive emotional responses, depending on the specific research goals. Furthermore, this approach enhances the development of more accurate and context-aware emotion recognition systems, with applications in mental health, virtual assistants, and customer service. By bridging the gap between semantic understanding and emotion detection, this research advances the state-of-the-art in SER and lays the groundwork for future exploration in the field.

\section*{Limitations}

This study, while providing valuable insights into the segmentation of speech for emotion recognition, has limitations. 
First, the dataset used in this research is curated from emotionally evocative movie clips, which, although varied, may not fully represent the broad diversity of real-world speech interactions. The emotional expressions captured in these clips might not encompass the full spectrum of spontaneous and everyday speech, which could limit the generalizability of the findings.

Second, although we included baseline speech-based emotion classifiers, their performance was substantially lower than that of text-based models. This gap likely stems from the emotional neutrality of the participants’ tone and the nature of the task. However, future work could explore whether jointly modeling text and acoustic features, perhaps guided by semantic segmentation, might uncover latent prosodic patterns aligned with specific semantic roles.

Third, while the study distinguishes between descriptive and expressive semantics, it focuses primarily on self-reported emotional responses, which can be subjective and influenced by individual differences in emotional expression and perception. This subjectivity introduces variability in the emotional ratings, potentially affecting the accuracy and robustness of the regression models.

%These limitations highlight areas where future research can build upon the findings, refining the methodologies and expanding the scope of emotion recognition to include a wider variety of speech contexts and real-world applications.

\section*{Ethics Statement}

This research was approved by the Research Ethics Board of University of Ottawa. All participants provided their informed consent prior to participating in the study.  Participants had the option to withdraw from the experiment at any time and for any reason, including emotional distress. Data collected during the study were handled securely and used exclusively for research purposes.  All personal data was anonymized. %In addition,the research team has taken steps to ensure fairness in the data analysis process, avoiding biases in data collection, segmentation, and prediction. 

In Ethics Sheet for Automatic Emotion Recognition and Sentiment Analysis, ~\citet{mohammad2022ethics} provides a structured ethical framework for developing and deploying Automatic Emotion Recognition (AER) systems, along 50 ethical considerations. He specifically emphasizes on the risks of privacy violations, reinforcing biases, and potential misuse in surveillance or manipulation. This Ethics Sheet serves as a guide for responsible AER development, and encourages researchers to question \textit{why they automate, whose interests are served, and how success is measured}. 

Recognizing the ethical risks and potential misuse of SER technologies, we strongly caution against issues such as biases in emotion datasets, AI models enforcing rigid norms on emotional expression, and the exclusion of neurodiverse and marginalized groups. These concerns must be carefully addressed before deploying SER systems in real-world applications. We urge industries to adopt responsible, explainable, and inclusive AI development practices, ensuring that these technologies are fair, transparent, and beneficial to all users.

\section*{Acknowledgments}
We extend our gratitude to OrbMedic Inc. (Sonny Chaiwala), Ottawa, Canada, for their financial and technical support throughout this project. Their expertise and insights were pivotal in shaping the direction of our research and ensuring its successful execution. Additionally, we acknowledge the support provided by the Natural Sciences and Engineering Research Council of Canada (NSERC) and the Ontario Centres of Innovation (OCI).

\bibliography{custom}
\bibliographystyle{acl_natbib}

\appendix

\end{document}